\documentclass{article}

\usepackage{arxiv}
\usepackage[utf8]{inputenc} 
\usepackage[T1]{fontenc}    
\usepackage{hyperref}       
\usepackage{url}            
\usepackage{booktabs}       
\usepackage{nicefrac}       
\usepackage{float}
\usepackage{amsmath}
\usepackage{amssymb}
\usepackage{graphicx}
\usepackage{authblk}

\usepackage{caption}
\usepackage{subcaption}
\providecommand{\tabularnewline}{\\}
\floatstyle{ruled}
\newfloat{algorithm}{tbp}{loa}
\providecommand{\algorithmname}{Algorithm}
\floatname{algorithm}{\protect\algorithmname}

\usepackage{algpseudocode}

\usepackage{amsmath,amssymb}

\begin{document}

\title{Few-shot time series segmentation using prototype-defined infinite hidden Markov models}

\author[1]{Yazan Qarout}
\author[1]{Yordan P. Raykov}
\author[2,3]{Max A. Little}

\affil[1]{Department of Mathematics, Aston University, Birmingham, UK}
\affil[2]{Department of Computer Science, University of Birmingham, Birmingham, UK}
\affil[3]{Media Lab, Massachusetts Institute of Technology, Cambridge, MA 02139, USA}
\maketitle
\begin{abstract}
We propose a robust framework for interpretable,  few-shot analysis of non-stationary sequential data based on flexible graphical models to express the structured distribution of sequential events, using prototype radial basis function (RBF) neural network emissions. A motivational link is demonstrated between prototypical neural network architectures for few-shot learning and the proposed RBF network infinite hidden Markov model (RBF-iHMM). We show that RBF networks can be efficiently specified via prototypes allowing us to express complex nonstationary patterns, while hidden Markov models are used to infer principled high-level Markov dynamics. The utility of the framework is demonstrated on biomedical signal processing applications such as automated seizure detection from EEG data where RBF networks achieve state-of-the-art performance using a fraction of the data needed to train long-short-term memory variational autoencoders. 
\end{abstract}

\section{Introduction}
Machine learning applications to time series data often require combining flexible representations of complex high-dimensional signals with parsimonious structured modelling which facilitates problem understanding. Feature extraction techniques allow for better problem formulation through statistical or spectral data representations, enabling data-efficient model training. More recently, this is being achieved through data driven techniques using end-to-end Neural Network (NN) autoencoders for feature extraction \cite{vincent2010, lecun2015deep}, outperforming traditional feature engineering approaches across many domains. However, NN techniques do not work so well in limited data scenarios, or where model interpretability is a necessity. In sensitive applications such as health status monitoring, large amounts of training data are often unavailable, and the focus is on exploratory knowledge discovery rather than supervised classification. Many deep NN architectures struggle to incorporate specific domain knowledge or sequential characteristics which would allow for principled exploratory analysis and causal inference of the outcomes \cite{zhang2016understanding,goodfellow2014explaining}. These shortcomings of NN autoencoding have motivated a lot of work on the composition of generative models combining directed graphical models with flexible NN variational autoencoders \cite{johnson2016}. Two major challenges with this strategy are: (1) variational autoencoders require large amounts of training data; (2) we have limited control over the distribution of the latent space. In signal processing applications, time series VAEs still require large amounts of data to learn simple underlying features about the spectrum of the time series. Recurrent NNs tend to excel in capturing longer-term morphological dependencies, but are inefficient clustering even simple spectral charateristics. Explicit spectral representations are required in the NN architecture (i.e. such as \textit{spectral pooling} \cite{rippel2015spectral}) to achieve good information preservation of the spectrum and make them resilient in high signal-to-noise ratio problems. 
 
In order to address these challenges, we propose a flexible probabilistic modelling framework which embeds feature-free structure using prototype-based networks. Prototypical NNs \cite{snell2017prototypical} have become increasingly widely used for few-shot classification tasks where they work by mapping input data through some function (i.e. embedding specified using a NN) and leveraging the similarity of the embedding to some pre-specified class prototypes. The prototypes are often selected examples (or mean embeddings) from each class. Then, the output map is constructed from a softmax layer which uses the distance of inputs from the prototypes. This approach is equivalent to augmenting a standard feedforward NN architecture with a \textit{radial basis function} (RBF) network layer \cite{broomhead1988,poggio1990networks}, with fixed basis centers which are class prototypes. RBF networks have been widely used across many modelling problems, but less so as a building block in Bayesian hierarchical models or multilayer architectures. This can, most likely, be attributed to the difficulty of integrating RBFs into more complex models in a stable manner \cite{ yu2006design} and the difficulty of specifying the more complex non-linear activation parameters. RBF networks have been widely used for autoregressive modelling of time series data where they can capture flexible nonstationary distributions, however, they provide inefficient representations in the presence of discrete latent structure which leads to difficulties when selecting bases.

Discrete latent variable models such as the hidden Markov model (HMM) have been widely used to capture explicit time dependency in discrete signals, but struggle to leverage more complex state distributions which limits their interpretability and predictive performance. In this paper we propose a novel Bayesian nonparametric infinite hidden Markov model with emission distribution specified by Bayesian RBF networks (RBF-iHMM). The proposed RBF-iHMM model allows for interpretable decomposition of signals into an unknown number of non-stationary segments. Many conventional HMM-based techniques seek to decompose a signal into a Markov chain of linear stationary series \cite{fox2011sticky} and flexible nonparametric iHMMs often lead to \textit{over-partitioning} of the observed dynamics of interest, due to the limited expressive power of the emission model. Our proposed model can be seen as an extension of Markov-switching AR models \cite{fox2009}, in which the linear AR states are replaced with prototype-defined, nonlinear AR components.

\section{Preliminaries}
\subsection{Nonlinear autoregressive processes}

An order $r$ vector autoregressive (VAR) model is a random process which describes a sequence $(\mathbf{y}_1,\dots,\mathbf{y}_T) \in R^D$ as a function of previous values in the sequence and a stochastic term:
\begin{equation}
\mathbf{y}_{t} = \mathbf{\varphi}\left(\mathbf{y}_{t-1},\mathbf{y}_{t-2},\dots, \mathbf{y}_{t-r}\right) + \mathbf{\epsilon}_{t}\\
\end{equation}
where $r$ refers to the assumed time lag, $\mathbf{\varphi}: R^{D\times r} \to R^D$ is some function and $\mathbf{\epsilon}_{t}$ is a random noise at time $t$ which will model as a standard Gaussian. This model is a specialization of general state-space models where autoregressive dynamics are captured in the latent space, allowing for the observed space to be smooth, or for it to follow additional data generating processes \cite{ghahramani1996parameter,roberts2004general}. VAR processes are commonly used as a parametric model of the Power Spectral Density (PSD) of a time series, where the lag parameter $r$ controls the resolution of the spectral model. In the special case of linear, one-dimensional $\mathbf{\varphi}$ such that $\mathbf{\varphi}\left(y_t\right) = \sum_{n=t-r}^{t-1}a_ny_n$ with coefficients $\{a_1\dots a_r\}$ the PSD of the process is:
\begin{equation}
S\left(f\right)=\frac{\sigma^{2}}{\left|1-\sum_{j=1}^{r}A_{j}\exp\left(-i2\pi fj\right)\right|^{2}}
\end{equation}
where $S$ denotes the PSD at frequencies $f$ and $i$ denotes the imaginary unit. This means that the number
of non-zero AR coefficients determines the level of detail of the PSD which the  model can represent: there is a peak in the PSD for each complex-conjugate pair of roots of the coefficient polynomial \cite{little2019machine}. When it comes to modelling stationary non-Gaussianity,  nonlinear $\mathbf{\varphi}$ should be used, where flexible and scalable VARs have been defined using a \textit{radial basis function network} as a model for $\mathbf{\varphi}$. In a single layer setting, this leads to the process:
\begin{equation}
\mathbf{y}_{t} = \sum_{j=1}^J w_j\phi\left(d\left(\overline{\mathbf{y}} - \mathbf{c}_j\right)\right) + \mathbf{\epsilon}_t
\label{eq:RBF_layer_1}
\end{equation}
where $J$ is the number of neurons in the hidden layer, W = \{$w_1,\dots,w_J\}$ are the inferable weight parameters, $d(.)$ is any valid distance metric, $\overline{\mathbf{y}}_t = \left\{\mathbf{y}_{t-1},\dots,\mathbf{y}_{t-r}\right\}$, $\{\mathbf{c}_1,\dots,\mathbf{c}_J\}$ are hidden layer centers and $\phi$ is a pre-defined non-linearity such as a Gaussian function or \textit{polyharmonic spline} \cite{madych1990polyharmonic,beatson1997fast}. The RBF VAR defined above explicitly focuses on encoding the spectral characteristics of a signal and has been shown to provide a more efficient representation of periodic patterns, as compared to direct mapping of signal windows \cite{vesin1993, shi1999nonlinear}.

\subsection{Posterior properties of RBF networks}

Complex NNs have high capacity, but large NNs are hard to analyse directly and it is difficult to incorporate specific structural assumptions about the data, such as specific domain knowledge or relevant causal structure. This can be done implicitly in deep generative models, either through direct modelling of the network architectures, or by exploiting implicit inductive biases in the neural autoregressive models \cite{uria2016neural} and other normalizing flows \cite{dinh2014nice,papamakarios2019normalizing}.  
By contrast, RBF networks can be intuitively analysed in the Bayesian setting and they can be used to explicitly control the induced probability distributions on the outputs, with appropriate modelling over the network parameters. In the single layer feed-forward setup, this will allow us to express the distribution induced from the inputs to the outputs of the network. For a single layer feed forward architecture $\varphi\left(\cdot\right)$, if we assume a zero-mean Gaussian over the weights $w\sim\mathcal{N}\left(0,\sigma_w^2\right)$ and zero-mean bias $\mathbf{\epsilon}_t \sim \mathcal{N}\left(\mathbf{0}, \sigma_{\epsilon}^2\right)$, we can compute the expectations with respect to the network weights $w$ over the random process defined by the network outputs: 
\begin{equation}
\begin{aligned}
\mathbb{E}_w\left[\varphi\left(\mathbf{y}\right)\right] & = \mathbf{0}\\
\mathbb{E}_w\left[\varphi\left(\mathbf{y}\right)\varphi\left(\mathbf{y}'\right)\right] & = \sigma_\epsilon^2 + J\sigma_w^2\mathbb{E}_c\left[\phi(\mathbf{y},\mathbf{c})\phi(\mathbf{y}',\mathbf{c})\right]
\end{aligned}
\end{equation}
with $J$ being the number of neurons and i.i.d. basis centers $\left\{c_1,\dots,c_J\right\}$ \cite{Williams1997computing}. As the number of neurons $J\to\infty$, \cite{neal1996bayesian} has shown that the distribution over the outputs of such a feed-forward network converges to a (stationary) Gaussian process at a rate of $J^{-1/2}$. 

Let us consider the widely-used special case where $\phi\left(\cdot\right)$ is a Gaussian basis function and $d\left(\cdot\right)$ is the Euclidean distance, i.e. $\varphi(\mathbf{y}) = \sum_{j=1}^J w_j\exp\left(-\frac{\lVert \mathbf{y} - \mathbf{c}_j\rVert}{\eta}\right) + \mathbf{\epsilon}_t$. For Gaussian basis centers $\left\{c_1,\dots,c_J\right\} \sim \mathcal{N}\left(0,\sigma_c^2\right)$, $\eta$ and $\sigma_c^2$ define the radius over which the expected value of the process will be significantly different from zero. With increasing basis center variance $\sigma_c^2$, we need an increasing number of hidden layers to achieve a comparable  quality of approximation. However, for finite $J$, the covariance of the stochastic process describing the outputs of the RBF network $\varphi$ is only locally stationary due to the lack of translation invariance over the weights $W$ \cite{Williams1997computing,meronen2020stationary}. The composite covariance function consists of a stationary covariance modulated with a Gaussian decay envelope:
\begin{equation}
\begin{aligned}
\kappa_{\text{RBF-NN}}\left(\mathbf{y},\mathbf{y}'\right) \propto \exp\left\{-\frac{\mathbf{y}^T\mathbf{y}}{2\tilde{\sigma}^2}\right\}\times\exp\left\{-\frac{(\mathbf{y}-\mathbf{y}')^T(\mathbf{y} - \mathbf{y}')}{2\hat{\sigma}^2}\right\}\times\exp\left\{-\frac{\mathbf{y}'^T\mathbf{y}'}{2\tilde{\sigma}^2}\right\} 
\end{aligned}
\end{equation}
with envelope and kernel variances, $\tilde{\sigma}^2$ and $\hat{\sigma}^2$ being functions of $\sigma_c^2$ and $\eta$: $\tilde{\sigma}^2 = 2\sigma_c^2 + \eta$ and $\hat{\sigma}^2 = 2\eta  + \eta^2/\sigma_c^2$ for the specific Bayesian RBF network setup as shown in \cite{Williams1997computing}. Similar results can be obtained for a larger set of nonlinearities $\phi$ which belong to the Matern class for which we can write the composite covariance function over the network outputs in the form:

\begin{equation}
\begin{aligned}
\kappa_{\text{RBF-NN}}\left(\mathbf{y},\mathbf{y}'\right) \propto \exp\left\{-\frac{\mathbf{y}^T\mathbf{y}}{2\tilde{\sigma}^2}\right\} \times\kappa_{\text{Matern}}\left(\mathbf{y},\mathbf{y}'\right)\times\exp\left\{-\frac{\mathbf{y}'^T\mathbf{y}'}{2\tilde{\sigma}^2}\right\} 
\end{aligned}
\end{equation}
with $\kappa_{\text{Matern}}\left(\cdot,\cdot\right)$ denoting the Matern class kernel matrix determined by the choice of basis $\phi$. A clear consequence is that the predictive likelihood over $\mathbf{y}_t$ (under a Bayesian RBF-NN) will behave like a predictive likelihood given a Gaussian process model for $\mathbf{\varphi}$ when the basis variance and length-scales, $\sigma_c^2$ and $\eta$, are sufficiently large. \cite{broomhead1988} first showed that the network output for isotropic Gaussian kernel is a linear combination of the radial basis functions.

\begin{figure}[!h]
\centering
\includegraphics[scale=0.35]{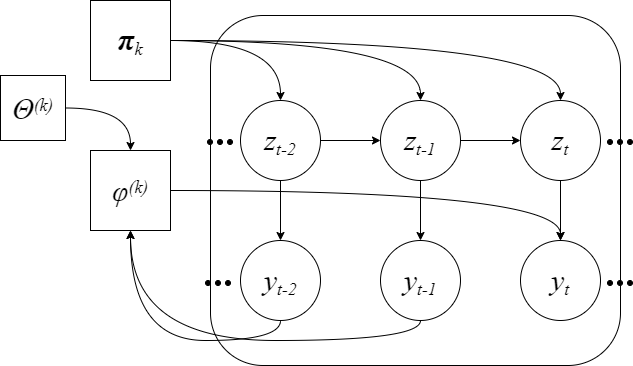}
\caption{Probabilistic graphical model of the RBF-iHMM. The latent state indicator variables $\{z_{t}\}$ are characterised by the transition distribution $\pmb{\pi}_{k}$. Given $z_{t}$ and the autoregressive observations $\{y_{t-1}, y_{t-2}, \dots\}$, $y_{t}$ is sampled from the RBF model $\varphi_{k}$ which is characterised by parameters $\mathbf{\Theta}$.}
\label{fig:pgm} 
\end{figure}
\section{Proposed method}

\subsection{Problem formulation}
Many problems in signal processing and time series analysis can be viewed as some kind of segmentation problem in which we wish to separate, or group observed states and study the process of their occurrence, and factors that affect these processes. This problem structure can be naturally formulated using discrete latent variable graphical models, such as the HMM. By fitting HMMs, we can discover natural clusters of patterns or activities occurring in time, the most likely states that precede these patterns, as well as estimate the uncertainty associated with an inferred segmentation. Such explicit discrete structures are difficult to represent directly using ``black-box'' NNs alone. The trade-off is that parametric discrete models such as HMMs can be quite rigid, by fixing the dimensionality of the latent state space. To allow for more flexible learning of the state space, we can use Bayesian nonparametric HMMs where the number of states is not fixed a priori. This facilitates better discovery of states as more data is presented, but is also known to lead to potential over-partitioning \cite{fox2011sticky, miller2013simple} of the data, which damages the interpretability of the inferred segmentation. This problem is easily seen when trying to represent non-Gaussian data with a mixture of Gaussian densities \cite{johnson2016} or non-stationary time signals with combination of linear ARs. In this work we try to address this problem by composing more flexible, non-linear network-defined state models, see Figure \ref{fig:pgm}.

\subsection{Model specification}
\begin{algorithm} 
   \caption{Inference of the novel RBF-iHMM model}
    \begin{algorithmic}[1]
    \label{algo}
	\item Initialise the transition distribution $\Pi$, emission distribution parameters $\{W^{(k)}\}$, $\{\mathbf{c}^{(k)}\}$ and truncation level $L$.
	
    \item Compute the HMM backward message $m_{t,t-1}$: 
    
    \begin{equation*}
    \begin{aligned}
    m_{t,t-1}(k)= \\
    \sum^{L}_{q=1}\pi_{k,q} \mathcal{N}&\left(\mathbf{\varphi}(\overline{\mathbf{y}}_{t}; W^{(k)}, \mathbf{c}^{(k)}), \Sigma^{(k)} \right)m_{t+1,t}(q)
    \end{aligned}
    \end{equation*}

    \item Compute the forward message for each time step $l(\mathbf{y}_{t})$ given the backward message $m_{t,t-1}$ .
    
    \begin{equation*}
    l_{k}(\mathbf{y}_{t}) = \mathcal{N}\left(\mathbf{\varphi}(\overline{\mathbf{y}}_{t}; W^{(k)}, \mathbf{c}^{(k)}), \Sigma^{(k)} \right) m_{t+1,t}(k).
    \end{equation*}

    \item Sample the indicator variables $(z_{1}, \dots, z_{T})$ 
    	\begin{equation*}
z_{t} \sim \sum_{k=1}^{L} l_{k}(\mathbf{y}_{t}) \pi_{z_{t-1},k} \delta(z_{t},k).
	\end{equation*}
where $\delta\left(\cdot\right)$ is the Kronecker delta.

    \item Sample the transition matrix $\mathbf{\pi}$ given mixing parameters $\mathbf{\beta}$ and the transition counts $n$:
    
    \begin{equation*}
    \begin{aligned}
    \mathbf{\pi}_{k} \sim Dir \left(\alpha \beta_{1} +n_{k,1}, \dots,  \alpha \beta_{k} +\lambda + n_{k,k} \right)
        \end{aligned}
    \end{equation*}
    for $k = 1,\dots, L$ and $\lambda$ being a ``sticky-state'' parameter enforcing self-transitions. 
    \item Sample the top level mixing parameters $\beta$.
    \item Update the radial basis centres $\mathbf{c}^{(k)}$:
    
    \begin{equation*}
        \mathbf{c}_j^{(k)} \sim \mathcal{N}(\overline{\mathbf{y}}^{(k)}, B^{(k)}),
    \end{equation*}
    for all neurons $j = 1,\dots,J$ and states $k = 1,\dots,L$ where $\overline{\mathbf{y}}^{(k)}$ denotes the sufficient statistics of the lagged observations associated with state $k$.
    \item Update the emission distribution parameters $\Sigma^{(k)}$ and $W^{(k)}$ given the prioir prameters $n_0$ and $S_0$, state observation count $N^{(k)}$ and the posterior parameters for each state $k$: $S^{(k)}_{\Phi \Phi}, S^{(k)}_{Y \Phi}, S^{(k)}_{YY}$ and $S^{(k)}_{Y} \vert \Phi$ (please see the appendix for derivation):
    
    \begin{equation*}
    \begin{aligned}
    \Sigma^{(k)} & \sim \mathcal{IW} \left(n_{0} + N^{(k)}, S^{(k)}_{Y} \vert \Phi +S_{0} \right),\\
    \Theta^{(k)}  \vert \Sigma^{(k)} & \sim \mathcal{MN} \left(S^{(k)}_{Y\Phi} \frac{1}{S^{(k)}_{\Phi \Phi}}, \Sigma^{(k)}, S^{(k)}_{\Phi \Phi} \right). 
    \end{aligned}
    \end{equation*}

\end{algorithmic}
\end{algorithm}

An HMM expresses a  probability distribution over a sequential data $Y = (\mathbf{y}_{1}, \mathbf{y}_{2}, \dots , \mathbf{y}_{T})$ by invoking a Markov chain of discrete, hidden, state variables $\mathbf{z} = (z_{1}, \dots, z_{T})$. The Markov dynamics is parametrized by the transition matrix $\Pi = \{\pmb{\pi}_1, \dots, \pmb{\pi}_K\}$ such that: $\pmb{\pi}_{k} = \{\pi_{ki}\}$ and
$\pi_{ki} = P\left(z_t = k \vert{z_{t-1} = i} \right)$ where $k,i = 1,\dots,K$ and $K$ denotes the number of states in the HMM. Each row of $\Pi$ describes a categorical distribution with $z_{t-1}$ pointing to the row of $\Pi$ specifying the likelihood of $z_t$.

In HMM-based switching autoregressive models, we assume that $\mathbf{y}_t$ is modelled by one of the autoregressive maps $\mathbf{\varphi}^{(z_t)}\left(\overline{\mathbf{y}}_t\right)$ with probability modelled by the Markov dynamics over the latent $\mathbf{z}$'s; $\overline{\mathbf{y}}_t = \left\{\mathbf{y}_{t-1},\dots,\mathbf{y}_{t-r}\right\}$ for an order $r$ autoregressive process. This allows decomposition of the likelihood at time index $t$ as:
\begin{equation}
    p(\mathbf{y}_t\vert Y^{(-t)},\mathbf{z},\mathbf{\Theta})\propto p\left(\mathbf{y}_t\vert \mathbf{\varphi}^{(z_t)}\left(\overline{\mathbf{y}}_t\right)\right)p(z_t\vert z_{t-1})
\end{equation}
where  $Y^{(-t)} = \left\{\mathbf{y}_i: \forall i\neq t \right\}$.  If our goal is predicting future $\mathbf{y}_{t+1}$, many flexible models for $\mathbf{\varphi}\left(\cdot\right)$ can bypass the need for decomposition of the likelihood into different $\{\mathbf{\varphi}^{(1)}\left(\cdot\right),\dots,\mathbf{\varphi}^{(K)}\left(\cdot\right)\}$ maps and latent $\mathbf{z}'s$ altogether. However, the augmentation of discrete Markov variables in HMMs, allows us to make structural assumptions about the underlying process distribution and make direct causal queries about the observed series: i.e. queries about how changes in $\mathbf{\varphi}^{(z_{t-1})}\left(\cdot\right)$ affect $\mathbf{y}_t$ (i.e. subject to valid model assumptions). In principle, variable order latent dependence in time can be modelled by augmenting the transition matrices \cite{DuPreez1998}, however for longer-term dependencies this raises issues related to \textit{credit diffusion} \cite{bengio1995diffusion} and additional modelling assumptions are required \cite{wood2011sequence, dedieu2019}.

Motivated by continuous monitoring applications, where the number of observed states $K$ grows with the volume of observed data, we specify a Bayesian nonparametric prior over $\Pi$ which implies $K\to\infty$ and the number of states represented in the observed data, $K^+$, is unknown. This is done by specifying the distribution of the transition matrix $\Pi$ using hierarchical Dirichlet process (DP) priors \cite{teh}, including a reinforced probability of self-transitions \cite{fox2011sticky}. Given the latent $z_1,\dots,z_T$, we define a RBF network-based model over the state specific $\{\mathbf{\varphi}^{(1)}\left(\cdot\right),\dots,\mathbf{\varphi}^{(K^+)}\left(\cdot\right)\}$ maps. The likelihood for $\mathbf{y}_t$, given the latent states, then takes the form:
\begin{equation}
    p(\mathbf{y}_t\vert \overline{\mathbf{y}}_t,z_t,\mathbf{\Theta})\propto \mathcal{N}\left(\mathbf{y}_t \vert \mathbf{\varphi}\left( \overline{\mathbf{y}}_t;\Theta^{(z_t)}\right), \Sigma^{(z_t)}\right)
\end{equation}
where $\Theta^{(k)} = \{\mathbf{\theta}_1^{(k)},\dots,\mathbf{\theta}_J^{(k)}\}$ denotes the whole set of network parameters (i.e. for each neuron) associated with the $k$-th state-specific RBF network and $\Sigma^{(k)}$ is the state specific noise. The full model then can be written as:

\begin{equation} \label{hdp}
\begin{aligned}
G_{0} \vert \gamma, H & \sim DP(\gamma,H), \\
G_{k} \vert \alpha, G_{0} & \sim DP(\alpha, G_{0}), \\
z_{t} | z_{t-1} = k & \sim G_{k},\\
\mathbf{y}_{t} \vert z_{t}, \overline{\mathbf{y}}_t & \sim \mathcal{N}\left(\mathbf{\varphi}(\overline{\mathbf{y}}_t; \Theta^{(z_{t})}), \Sigma^{(z_{t})} \right),
\end{aligned}
\end{equation}

where $H$ and $\gamma$ are the base probability measure and concentration parameter of the upper level DP, $G_0$ and $\alpha$ are the base probability measure and concentration parameters of the lower level DP, and $G_{k}$ are a set of random probability measures specifying the state specific distribution. The base measure $G_k$ reflects the expected value of the state-specific models, where $H$ reflects the process generating the state-specific model parameters. In a standard Gaussian emission model or linear AR emission model a closed form conjugate prior for $H$ can be selected, however, in the proposed setup integrating over $\mathbf{\Theta}$ is rarely feasible and leads to slow Markov chain mixing during inference. Instead, $H$ specifies the prior belief over our network parameters $\mathbf{\Theta}$, which includes the network weights, basis centers and additional basis parameters depending on the RBF architecture $\mathbf{\varphi}\left(\cdot\right)$. 

We will focus on the single layer $\mathbf{\varphi}\left(\cdot\right)$, since it leads to readily intepretable nonlinear state-specific models, but multilayer extensions would be equivalent to placing a Baysian prototypical network on $\mathbf{\varphi}\left(\cdot\right)$. Then our model for $\mathbf{\varphi}\left(\cdot\right)$ takes the form of Equation \ref{eq:RBF_layer_1}. In a fully Bayesian setting, we can specify our base measure $H$ as a Gaussian describing the basis centers $\mathbf{c}_1^{(k)},\dots,\mathbf{c}_J^{(k)}$ for all states $k$, a set of independent Gaussians describing the weights $W^{(k)} = \{\omega_1^{(k)},\dots,\omega_J^{(k)}\}$ and for certain basis functions $\phi$ a prior for the nonlinearity shape parameter. A common basis function $\phi$ which does not require tune-able shape parameters is the \textit{polyharmonic spline} \cite{madych1990polyharmonic,beatson1997fast}. 

The  observation likelihood for instantiated state $k\in\{1,\dots,K^+\}$ can be written in the general form:

\begin{equation}
p(\mathbf{y}_{t} \vert z_{t} = k) = \frac{exp \left( - d \left( \mathbf{y}_{t}, W^{(k)} \Phi^{(k)} \right)  \right)}{ \sum_{i = 1}^{(k)} exp \left( - d \left( \mathbf{y}_{t}, W^{(i)} \Phi^{(i)} \right)  \right)},
\end{equation}

where $J$ denotes the number of hidden nodes, $W^{(k)}$ is a $(D \times J)$ matrix of weights associated with state $k$, $\Phi^{(k)} = \phi \left(d(\overline{\mathbf{y}}_t, \mathbf{c}^{(k)}) \right)$ is the matrix form of the predefined non-linearity representing the hidden layer for state $k$, with centres $\mathbf{c}^{(k)}$ and distance $d(\cdot)$ assumed Euclidean unless specified otherwise. In practice, we instantiate a larger set of networks with $\mathbf{\Theta} = \{\Theta^{(1)},\dots,\Theta{^{(L)}}\}$ for $L >> K^+$ to allow for data-driven learning of the represented $K^+$ by truncation. Completely unseen states can be also instantiated via the stick-breaking construction \cite{sethuraman1994constructive}, but for most basis and networks, non-conjugate methods are required.

\begin{figure*}[!h]
\begin{centering}
{\hspace{0.3cm} \Large $\Pi_{actual}$ \hspace{1.6cm} $\Pi_{RBF-iHMM}$ \hspace{0.9cm} $\Pi_{AR-iHMM}$} \\
\medskip{}
\includegraphics[scale=0.37]{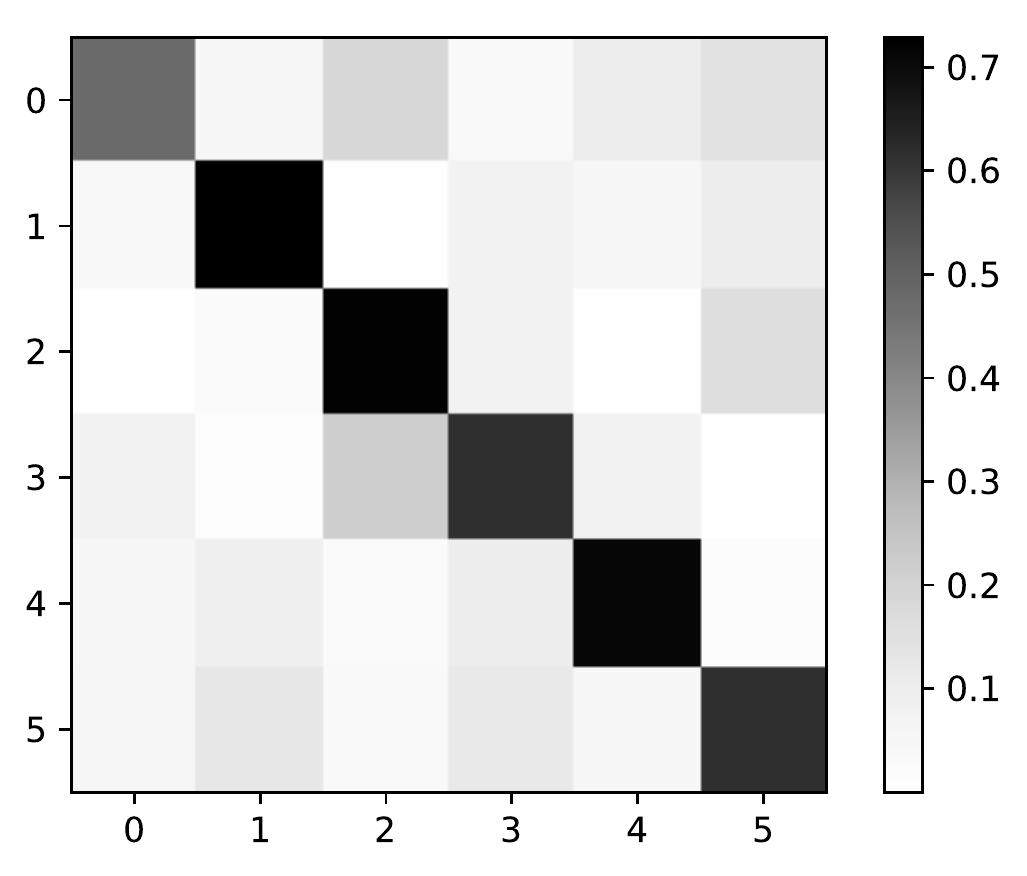}
\includegraphics[scale=0.37]{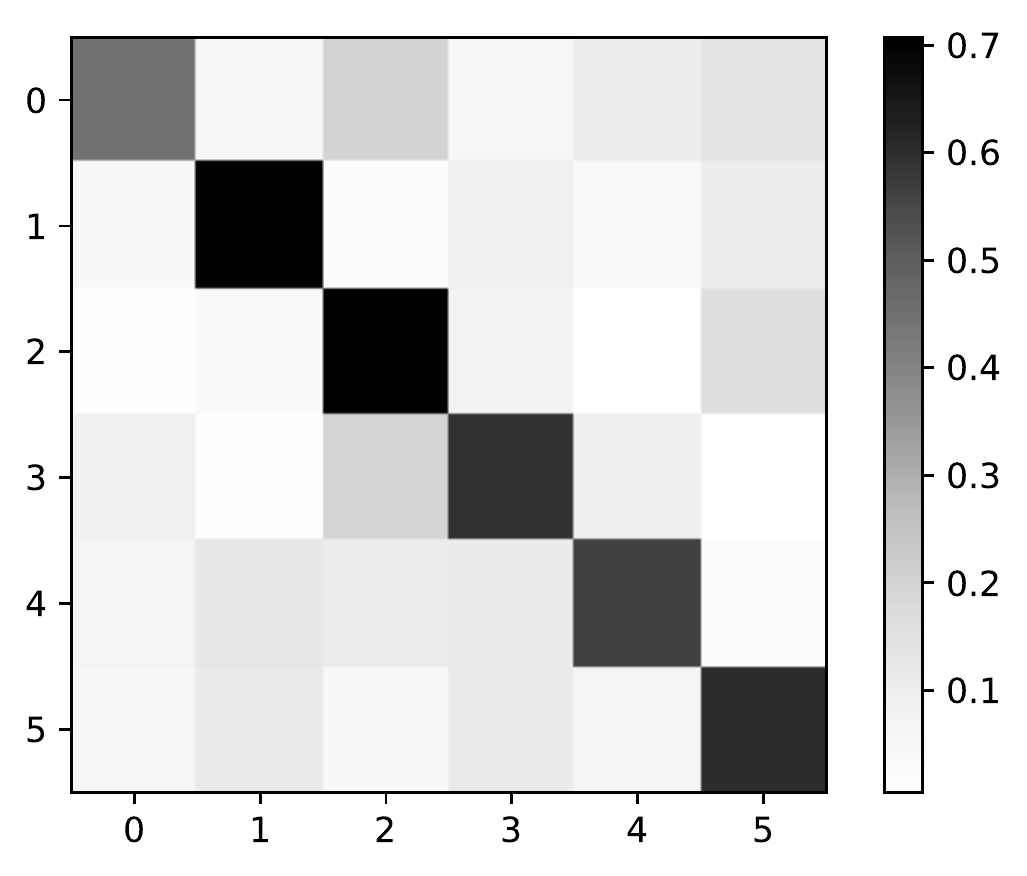}
\includegraphics[scale=0.37]{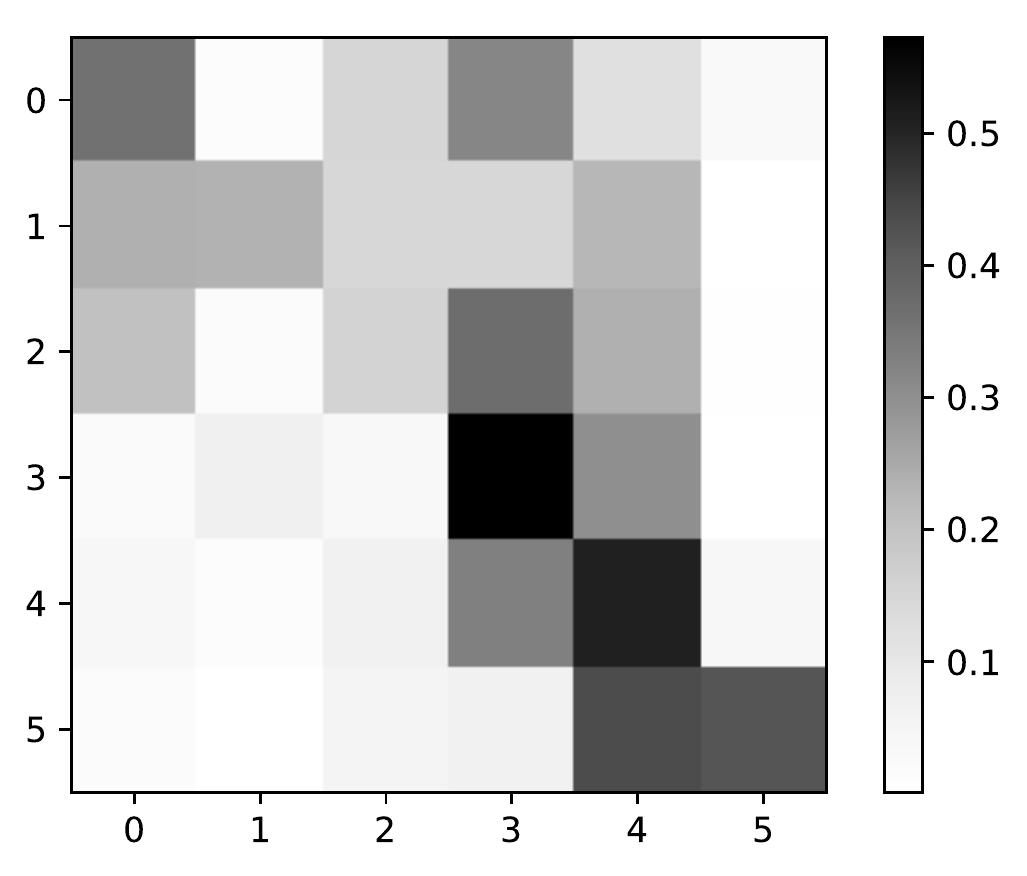}\\
\end{centering}
\caption{The respective matrices represent the actual transition matrix of the synthetic data along with the transition matrix as estimated by the RBF-iHMM and AR-iHMM models. The proposed non-linear switching RBF-iHMM was capable of estimating the true data generating transition matrix accurately, while the AR-iHMM with linear emissions failed.}
\label{fig:synth2} 
\end{figure*}

\subsection{Model optimisation}
The main challenge in combining structured graphical models with components specified by neural networks comes with the derivation of stable inference. One way to do this is using VAEs or approximate ``black-box'' variational inference \cite{ranganath2014black}. An advantage of RBF network-based states is that they allow for intuitive Bayesian treatment of the network parameters and closed form approximate inference. Training the network parameters in an unsupervised setup is ill-posed. This is why here we adopt a semi-supervised prototype-based strategy for selecting the state-specific network parameters: basis centers $\mathbf{c}_k$ are selected based on prototype examples (i.e. or their embeddings in a multilayer setup). We perturb the basis centers with additive Gaussian noise to get: $\mathbf{c} \sim \mathcal{N}\left(\overline{\mathbf{y}}, B \right)$. The sufficient statistics $\overline{\mathbf{y}}$ are selected to incorporate prior information gained by selected prototypes, such as the mean of lagged observations $\overline{\mathbf{y}}$ representative of class $l\in\{1,\dots,L\}$. 

In the RBF network literature, basis centers have been also set by a pre-clustering stage, where centers are placed at estimated cluster centroids \cite{chen1991orthogonal}, or through selection of diverse representatives of different class distributions \cite{kulesza2012determinantal}. However, in this work $\overline{\mathbf{y}}$ is simply the sample mean of prototypes of each class. The iHMM model then learns how many prototype classes are supported in the data.

For the special case of Euclidean $d\left(\cdot\right)$, single layer $\mathbf{\varphi}\left(\cdot\right)$ and fixed weights $W$ (i.e. without a Bayesian prior), the weights update given the state indicators and basis centers is a convex optimization problem which can be solved more efficiently; for more complex architectures we could use standard stochastic gradient descent or fully Bayeisan inference to learn the network parameters efficiently. Given the state specific network parameters, we infer the state indicators $\textbf{z}$ and the transition matrix using an efficient blocked Gibbs sampler which is equivalent to the update for linear iHMM AR models \cite{fox2009,qarout2020}.

\section{Experiments}
To demonstrate the effectiveness of our RBF-iHMM model, two experiments were conducted. First, we performed a synthetic study using generated piecewise non-stationary data where we demonstrate the flexibility of the RBF-iHMM as compared to a linear AR-iHMM counterpart. Our second experiment uses electroencephalogram (EEG) time series data from the UCI repository \cite{andrzejak2001} containing EEG time series with a sampling rate of 173.61Hz measuring patient brain activity at times when seizures were occurring and when symptoms were inactive. We compare the capacity of AR-RBF network components against VAE-LSTM classifiers in their ability to infer accurate class probabilities, when trained directly on raw EEG data. The two approaches are compared using different proportions of training data to demonstrate the capacity of the proposed prototypical RBF-iHMM components to encode informative features using very little training data. 
\begin{figure}[!h]
\centering
\includegraphics[scale=0.55]{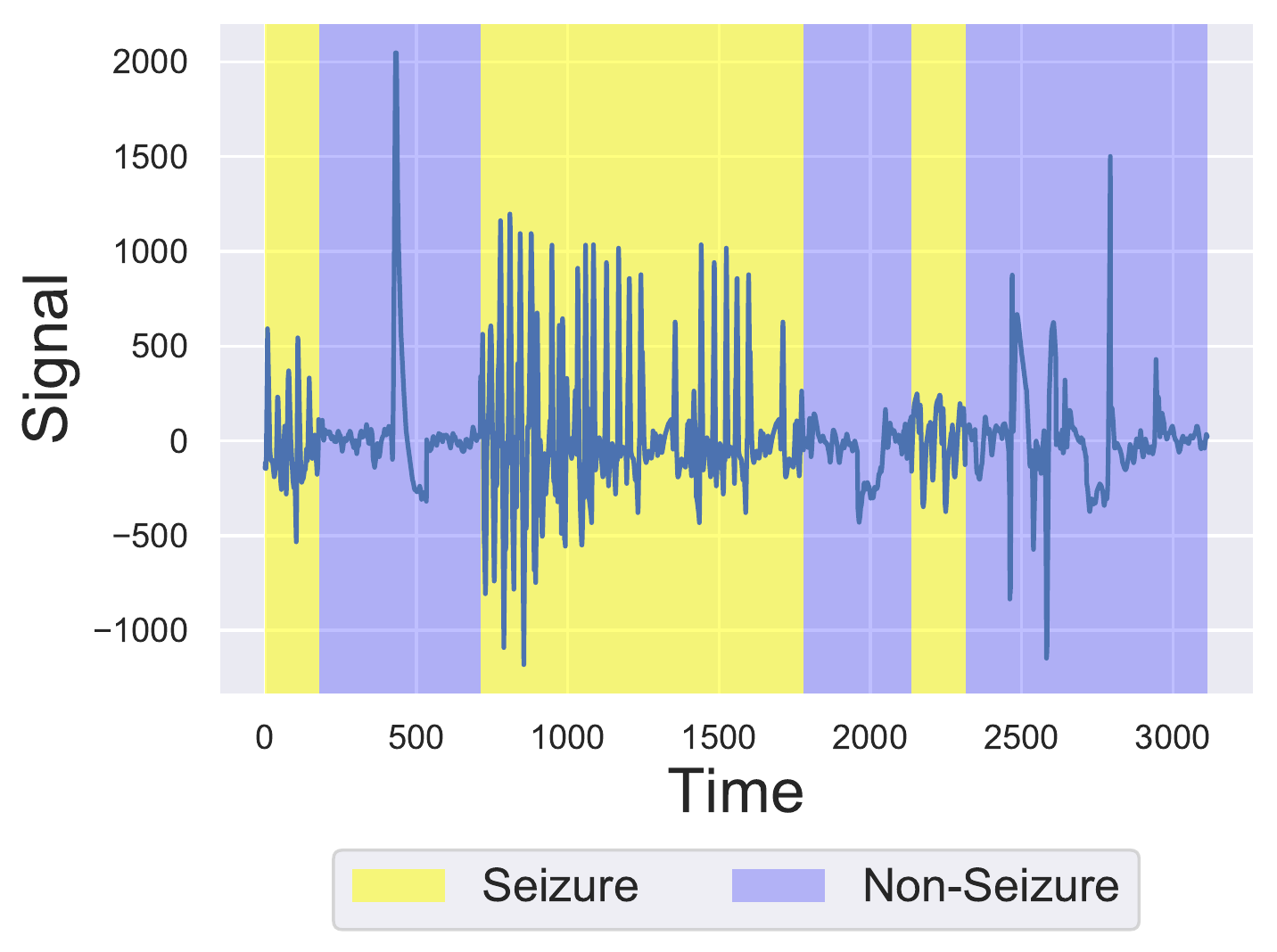}
\caption{A sample plot from the epileptic seizure recognition dataset depicting EEG signals for seizure and non-seizure conditions. The two states demonstrate piece-wise non-stationary patterns which will be over-represented with conventional linear emission models.}
\label{fig:signal} 
\end{figure}
\subsection{Synthetic data simulation}
Data was collected form six different switching nonlinear components which were generated using six AR RBF states sampled from a matrix normal distribution, enforcing no explicit sparsity on the weights of the network. The switching mechanism between the states followed a pre-defined transition matrix to yield a one-dimensional observation sequence $Y$ consisting of 10,000 time indices and a known latent state sequence $\textbf{z}$. 

The sequence $Y$ was modelled with both the RBF-iHMM and the AR-iHMM \cite{fox2009} with the goal of inferring the underlying generating transition matrix $\Pi_{actual}$, latent state sequence $\textbf{z}$ and the emission distribution parameters in an unsupervised approach. The HMM hyperparameters are identically fixed for both models, as well as the corresponding emission hyperparametrs. To compare performance, the estimation accuracy of $\textbf{z}$ and mean square error (MSE) of the transition matrix $\Pi$ entries were calculated (Table \ref{tsynth}). As it can be seen, the RBF-iHMM significantly outperforms the AR-iHMM with linear emissions and reproduces the underlying data generation mechanism accurately with a lower mean square error. This is presented visually in Figure \ref{fig:synth2} where the grayscale heatmaps of the predicted transition matrices of each technique are plotted alongside the actual underlying data generating transition matrix.


\begin{table}[htbp]
\centering \caption{Accuracy of the latent state variable $\textbf{z}$ and Mean square error for the estimation of the transition matrix $\Pi$ on the synthetic dataset using both the RBF-iHMM and the AR-iHMM respectively.}
\centering 
\begin{tabular}{lll}
\toprule 
  & RBF-HMM  & AR-HMM \tabularnewline
\midrule 
Prediction accuracy of $\textbf{z}$  & 0.95  & 0.53 \tabularnewline
MSE of $\Pi$  & $1.5\times10^{-4}$  & $1.1\times10^{-1}$ \tabularnewline
\bottomrule
\end{tabular}\label{tsynth} 
\end{table}


\subsection{Epileptic seizure recognition}

We demonstrate the capacity of the AR-RBF layers for few shot learning on EEG data where an experiment was setup on the Epileptic seizure recognition dataset \cite{andrzejak2001} of 100 participants. The data records the EEG activity of the participants at 5 different states including: eyes open, eyes closed, seizure episodes, EEG from tumor affected area of the brain and healthy brain EEG activity. For the purpose of this section, all non-seizure states were merged into one, yielding two states for classification: seizure and non-seizure. Seizure activity can be effectively detected with specialised, linear analysis approaches \cite{mcsharry2003prediction}. However for this example, we will be demonstrating non-linear state segmentation outperforming other non-linear and piecewise linear approaches. Moreover, the RBF-iHMM provides a generative model capable of seizure prediction whereas the techniques discussed in \cite{mcsharry2003prediction} can only be used for detection. Figure \ref{fig:signal} demonstrates a sample plot of the data where the non-stationary segments of the seizure/non-seizure trends can clearly be seen.

\begin{figure*}[!h]
\centering
\includegraphics[scale=0.5]{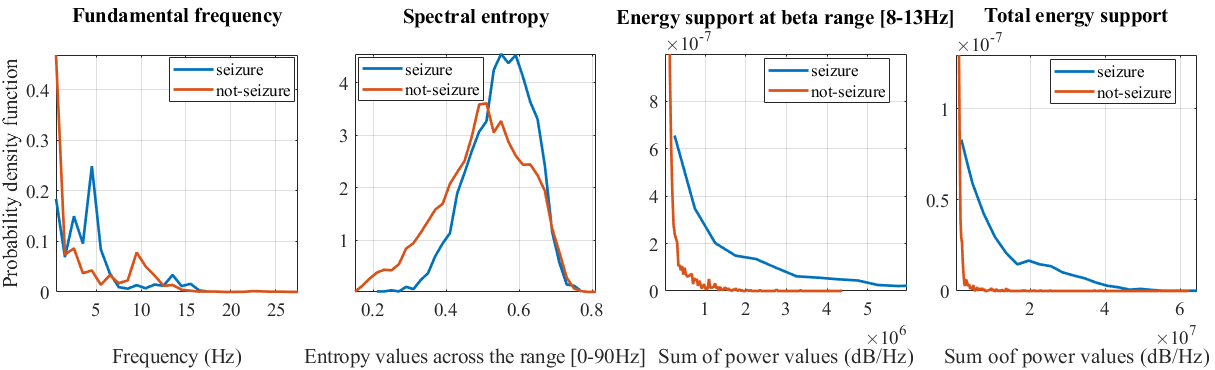}
\caption{Probability density functions of selected spectral features from the EEG data for seizure and non-seizure classes. The features are the fundamental frequency, spectral entropy and energy support on the alpha range [8-13Hz]  and they show moderate class separation.}
\label{fig:eegfreq} 
\end{figure*}

We first specify a distribution over the raw measurements for the two EEG classes (seizure/non-seizure) using two independent Bayesian AR-RBFs. Then, the observation class identity for new unseen data is calculated as the maximum a posteriori classification result (i.e. assigning each observation to the class with most likely embedding). This operation was completed using eight different training/test splits, where the training proportions were: 0.1\%, 0.3\%, 1\%, 5\%, 20\%, 40\%, 60\% and 80\%. Prior to the training/test splits, the data of 5 out of the 100 participants were completely removed from the dataset to use for validation and result demonstrations. The autoregressive order of the model was set to be a multiple of the seizure time period (26 data points) to reach slightly under one second's worth of data which amounts to six periods or 156 data indices. For comparison, the same experiments were conducted with a linear AR-iHMM components and a VAE-LSTM model with logistic regression classification at the embedding. The VAE-LSTM architecture \cite{chung2019unsupervised} consists of an encoder layer with 78 neurons followed by a fully connected 39 dimensional VAE layer, as well as two reverse decoder layers mapping from the latent layer to 78 dimensions then back to the original 156. This shallow VAE-LSTM model was selected to optimise the training efficacy under the small data conditions of this experiment, and for fair comparison against the RBF-iHMM's capability to deliver high accuracy performance with little training data. A deeper/more complex NN will require much more training. 

\begin{figure}[!h]
\centering
\includegraphics[scale=0.45]{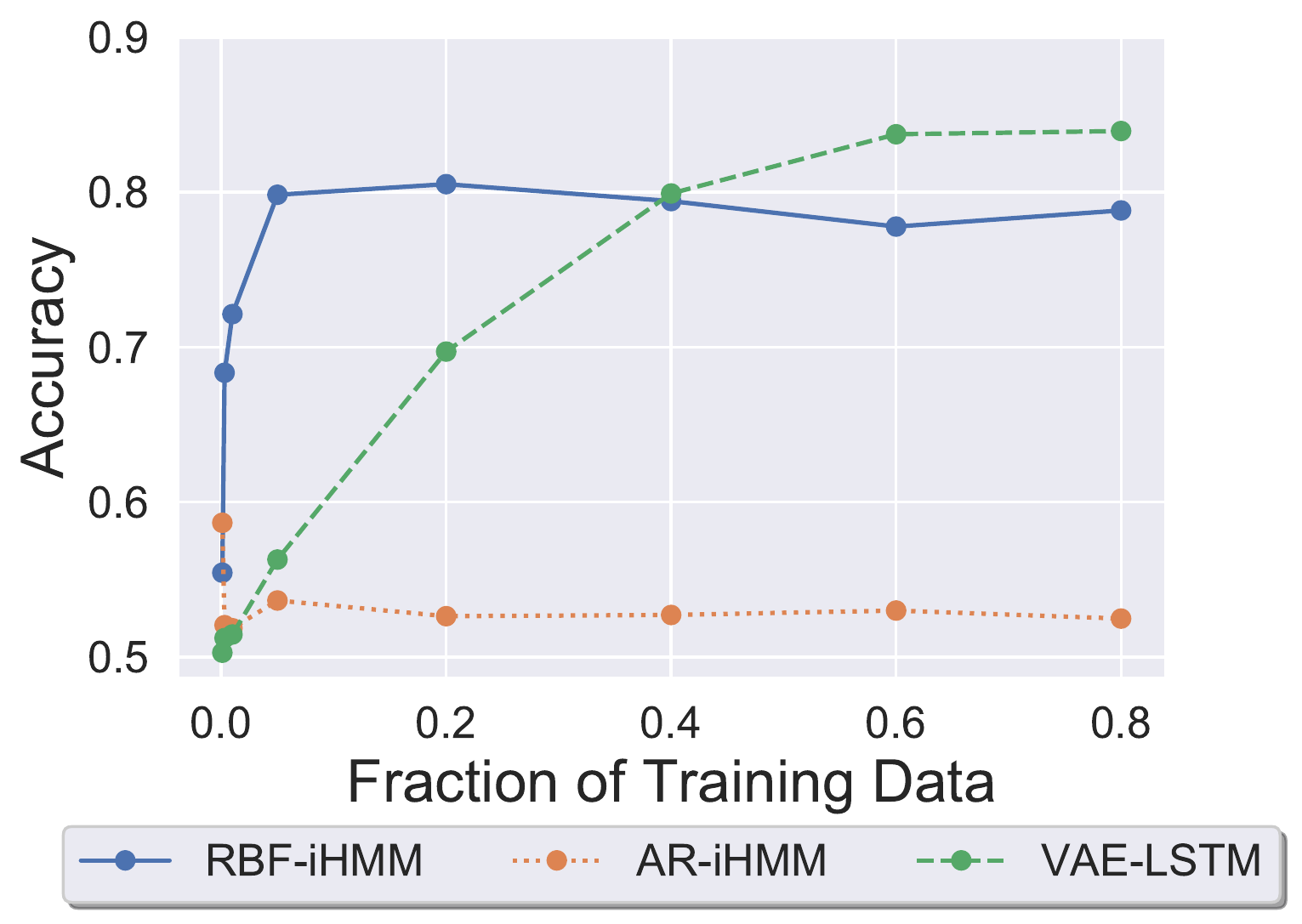}
\caption{Balanced accuracy for seizure/non-seizure classification obtained using the RBF-iHMM with Euclidean distance nonlinearities, linear emission AR-iHMM, and VAE-LSTM with logistic regression embedding classification when using different proportions of the data for training. The problem is most challenging at the left where we fit the models using $0.1\%$ of the data for training and $99.9\%$ for testing, to the right we use $80\%$ of the data for training and $20\%$ for testing.}
\label{fig:succ} 
\end{figure}

Experiments were repeated 20 times for each training/test split and the average performance accuracy on the validation set was plotted against the fraction of training data (split) in figure \ref{fig:succ}. As it can be seen, the RBF-iHMM components were capable of training effectively under few-shot training conditions due to the embedding of the frequency domain features through the autoregressive dissimilarity measure with the prototype radial basis centres. Figure \ref{fig:eegfreq} plots of some of the probability density functions of the frequency domain features extracted from the test set. It can be seen that there exists class distinction based on the data spectrum. The RBF-iHMM can provide high accuracy on test and validation with few training examples by optimising to encode these frequency domain features through appropriate prototype selection.
\begin{figure}[!h]
\centering
{\Large RBF-iHMM}\\
\includegraphics[scale=0.4]{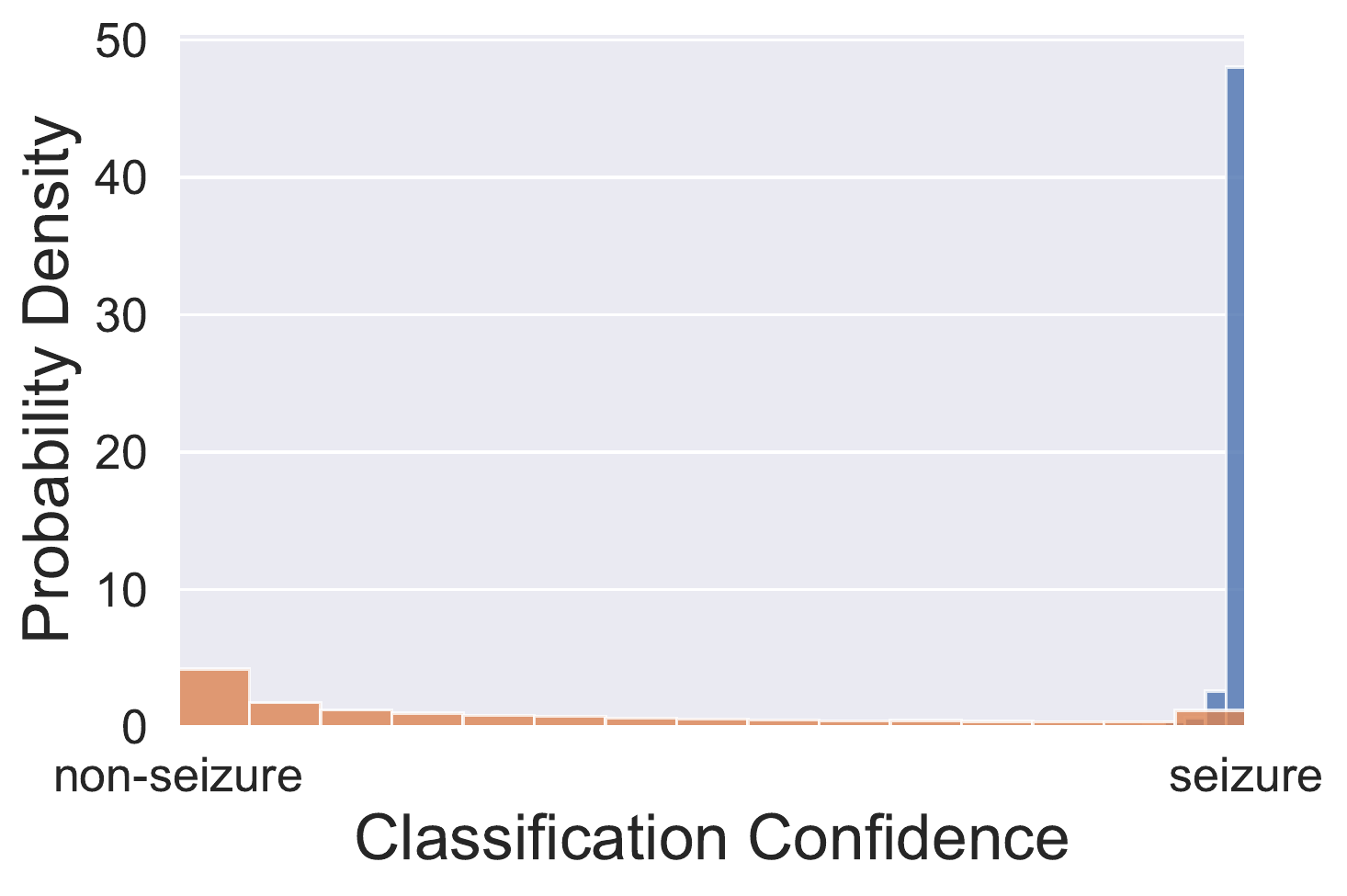}\\
\vspace{0.2cm}{\Large VAE-LSTM}\\
\includegraphics[scale=0.4]{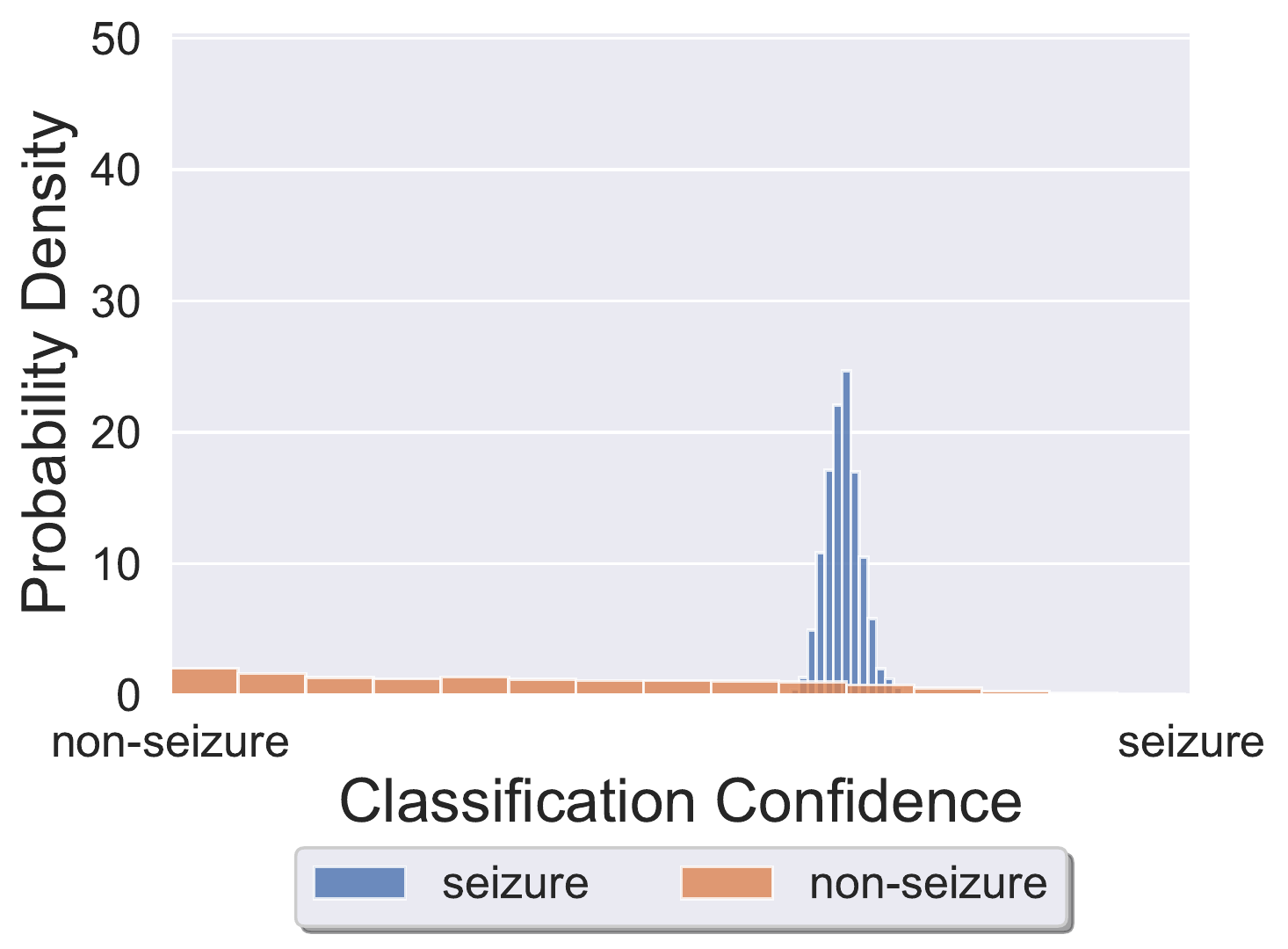}
\caption{The distribution of the classification confidences of the RBF-iHMM and the VAE-LSTM on the validation dataset using 20\% of data for training, where bars further to the right of the plot represent higher confidence in seizure classification, bars further to the left represent high confidence in non-seizure classification and the height of the bars represent the probability of an observation falling into this confidence level. Orange bars represent seizure observations and blue bars represent non-seizure.}
\label{fig:hist} 
\end{figure}
The proposed approach delivered an accuracy above 80\% using only 5\% of the data for training, compared with linear AR-iHMM components which did not capture the non-linear emission dynamics of the data resulting in an accuracy of approximately 50\% (random) on all training/test splits. The VAE-LSTM classifier showed a logarithmic increase in accuracy with the increase of training data. At 40\% split, the VAE-LSTM equals the performance of the RBF-iHMM and it eventually delivers higher accuracy when more training data is provided. However, this improvement is mostly due to the enhancement in capturing the non-seizure distribution which is a more difficult problem than modeling the seizure time series, since it includes multiple different types of EEG behaviours, whereas the seizure state contains only one kind of behaviour. The distribution of the classification confidence of the RBF-iHMM and the VAE-LSTM is visualised in Figure \ref{fig:hist}, where bars further to the right of the plot represent higher confidence in seizure classification, bars further to the left represent high confidence in non-seizure and the length of the bars represent the probability of an observation falling into this confidence level (orange bars for seizure observations and blue bars for non-seizure). It can be seen that the RBF-iHMM components capture the seizure distribution with very strong confidence, where there is a high probability that a new seizure observation will be clustered correctly. The VAE-LSTM also captures the state well but with lower confidence, new observations may be clustered as false negatives. By embedding the periodic feature space, the proposed RBF-iHMM seizure model component was capable of fitting the seizure distribution with high performance, if used independently as a classifier, model accuracy can be enhanced to 90\% by raising the classification threshold, outperforming even a fully trained VAE-LSTM.

\begin{table}[htbp]
\centering \caption{RBF-iHMM component performance accuracy using various distance metrics $d(\cdot)$ and number of centres $\pmb{c}$. Increasing the number of basis centres as a hyper-parameter improves the average model performance slowly reaching a plateau. The cosine dissimilarity metric proves to be more effective at embedding the frequency domain feature space than the standard euclidean distance by delivering accuracy higher by 2\%, while the Manhattan distance drops accuracy significantly by failing to effectively embed the feature space.}
\centering 
\begin{tabular}{lll}
\toprule 
 Emission model description & Accuracy  \tabularnewline
\midrule 
Euclidean distance with 10 basis centres & 0.60   \tabularnewline
Euclidean distance with 50 basis centres  & 0.67   \tabularnewline
Euclidean distance with 100 basis centres  & 0.74  \tabularnewline
Euclidean distance with 150 basis centres  & 0.75   \tabularnewline
Euclidean distance with 250 basis centres  & 0.81   \tabularnewline
Cosine dissimilarity with 250 basis centres  & 0.83   \tabularnewline
Manhattan distance with 250 basis centres  & 0.70   \tabularnewline
\bottomrule
\end{tabular}\label{tdist} 
\end{table}

The choice of the RBF-iHMM emission state hyperparameters can also influence the performance accuracy. Since, the model can train well with few training examples due to the prototype centre selection approach, the common risk of overfitting RBF networks is reduced. Therefore, using a large number of centres is possible, at the expense of computational time. The distance metric $d(\cdot)$ used in the basis function can also have significant effects on performance. For example, for the epileptic seizure dataset, the frequency domain features are important for classification, therefore, a distance metric that is efficient at encoding the spectrum is recommended. Table \ref{tdist} shows the RBF-iHMM component performance accuracy at 40\% training split using different model complexities and radial basis centre distance metrics. The results for the proposed technique demonstrated in Figure \ref{fig:succ} used a 250 centre model with Euclidean distance $d(\cdot)$. A larger number of basis centres leads to an asymptotic increase in the classification accuracy since this increases the probability of capturing the optimal embedded feature distribution through the selected prototypes. Euclidean distance is a suitable metric of capturing the frequency domain outperforming measures such as the Manhattan distance. However, the cosine dissimilarity captures the Fourier transform with higher resolution, delivering better feature embedding and therefore better average accuracy.


\section{Conclusion}
\label{sec:refs}

In this paper, we introduced the novel RBF-iHMM for few-shot, prototype-based learning of time series data with non-stationarity. The model combines inverse logic with an interpretable, non-parametric probabilistic structure for capturing non-stationary, and time-evolving emission states with high performance on out of sample data. Through experiments, we have shown that this new model performs with state-of-the-art accuracy, outperforming VAE-LSTMs, even with very few training examples by embedding the feature space of the raw data through prototype training examples. The model was presented with shallow emission architectures, however, it can be expanded to deeper and more complex models only at the expense of interpretability.

\newpage
\bibliographystyle{IEEEtran}
\bibliography{references,refs}
\end{document}